\documentclass[letterpaper, 10 pt, conference]{ieeeconf}  % Comment this line out if you need a4paper

\IEEEoverridecommandlockouts                              % This command is only needed if 
\usepackage{amsmath,amssymb,amsfonts}
\usepackage{algorithm}
\usepackage{algpseudocode} 
\usepackage{capt-of} 
\usepackage{graphicx}
\usepackage{textcomp}
\usepackage{xcolor}
\usepackage[dvipsnames]{xcolor}
\usepackage{stfloats}
\usepackage[backend=biber,style=ieee]{biblatex}
\usepackage{booktabs}
\usepackage{stfloats} % For better float handling
\usepackage{capt-of} 
\def\BibTeX{{\rm B\kern-.05em{\sc i\kern-.025em b}\kern-.08em
    T\kern-.1667em\lower.7ex\hbox{E}\kern-.125emX}}
\usepackage[bookmarks=true]{hyperref}

\title{\LARGE \bf
From Fold to Function: Simulation-Driven Design of Origami Mechanisms
}

\author{% <-this % stops a space
Tianhui Han, Shashwat Singh, Sarvesh Patil, Zeynep Temel\\
\small Robotics Institute, Carnegie Mellon University, Pittsburgh, PA 15213, USA
}

\begin{document}

% \maketitle
\thispagestyle{empty}
\pagestyle{empty}

\makeatletter
\let\old@maketitle\@maketitle
\renewcommand{\@maketitle}{%
  \old@maketitle
  % teaser figure spanning both columns on page 1
  \vspace{-0.5em}
  \begin{center}
    \includegraphics[width=\linewidth]{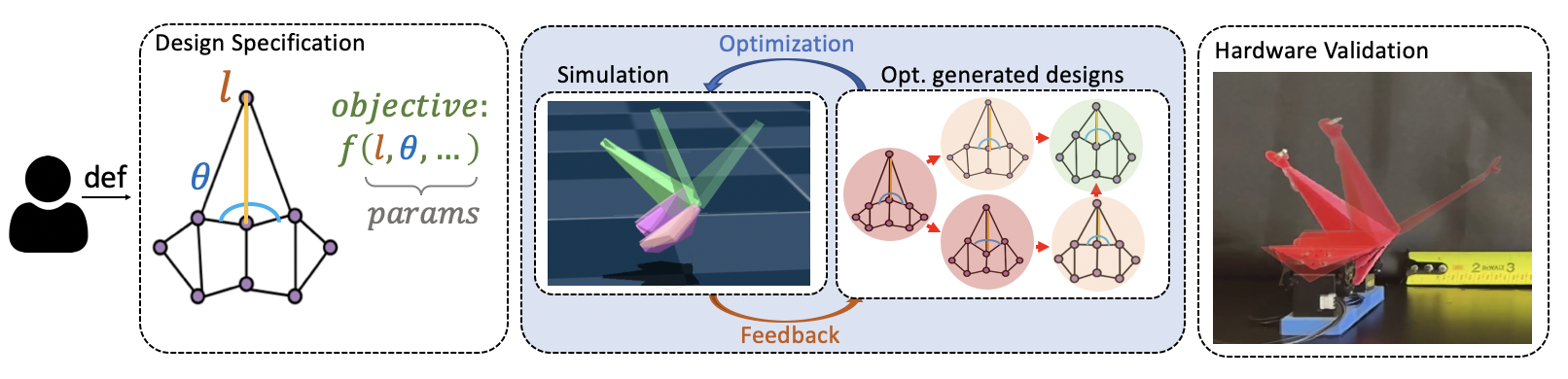}
    \captionof{figure}{An overview of the proposed framework: users define their origami designs through a graphical user interface (GUI), which automatically converts the specifications into a MuJoCo compatible simulation model. The simulation environment then generates data through dynamically interacting with the modeled mechanism which can be used as feedback for design optimization. We validate the final optimized configuration by fabricating the design in hardware and demonstrating improved performance compared to non-optimized variants.}
    \setcounter{figure}{1} 
    \label{fig:overall_fig}
  \end{center}
  \vspace{-1em}
}
\makeatother
\maketitle
%%%%%%%%%%%%%%%%%%%%%%%%%%%%%%%%%%%%%%%%%%%%%%%%%%%%%%%%%%%%%%%%%%%%%%%%%%%%%%%%
\begin{abstract}
Origami-inspired mechanisms can transform flat sheets into functional three-dimensional dynamic structures that are lightweight, compact, and capable of complex motion. These properties make origami increasingly valuable in robotic and deployable systems. However, accurately simulating their folding behavior and interactions with the environment remains a challenge. To address this, we present a design framework for origami mechanism simulation that utilizes MuJoCo’s deformable body capabilities. In our approach, origami sheets are represented as graphs of interconnected deformable elements with user-specified constraints such as creases and actuation, defined through an intuitive graphical user interface (GUI). This framework allows users to generate physically consistent simulations that capture both the geometric structure of origami mechanisms and their interactions with external objects and surfaces. We demonstrate our method's utility through a case study on an origami catapult, where design parameters are optimized in simulation using the Covariance Matrix Adaptation Evolution Strategy (CMA-ES) and validated experimentally on physical prototypes. The optimized structure achieves improved throwing performance, illustrating how our system enables simulation-driven origami design, optimization, and analysis.
% Origami-inspired robots leverage folding 2D surfaces to achieve complex 3D transformations with targeted actuation. Having a simulation platform that could allow designers to quickly iterate and optimize their origami mechanisms. However, having a platform that could simulate the interaction of mechanisms with other objects and environments could be challenging. In this paper, we represent pre-folded origami sheets as a graph of interconnected deformable elements with user-specified constraints such as creases and actuation definable through an intuitive Graphical User Interface (GUI). Based on the definition, we introduce a design framework for origami mechanism simulation utilizing MuJoCo's deformable structures capabilities. MuJoCo's deformable simulation support allows us to reliably capture implicit compliance of deformable sheets as well as the dynamic interactions of the mechanisms with other environment such as rigid objects and surfaces. We then demonstrate how our framework can be leveraged for design analysis by optimizing an origami catapult mechanism in simulation and validating in hardware that the optimized configuration outperforms non-optimized variants in a physical experiment, thereby reducing the need for repeated physical prototyping.  
% \textcolor{red}{Since deadline was extended could we do final feedback after fall break? Thanks!}
\end{abstract}

%%%%%%%%%%%%%%%%%%%%%%%%%%%%%%%%%%%%%%%%%%%%%%%%%%%%%%%%%%%%%%%%%%%%%%%%%%%%%%%%
\section{Introduction}

Origami-inspired design has evolved from an ancient Japanese artistic practice to a powerful engineering design paradigm~\cite{turner2016review}. By designing complex motion into a flat sheet of material, origami principles offer exceptional scalability and manufacturing efficiency, allowing devices to be shipped in a flat or folded form and deployed on demand~\cite{rus2018design}. Building on these advantages, origami mechanisms enable compact, lightweight, and reconfigurable systems in diverse domains, such as biomedical devices~\cite{gultepe2012biopsy, johnson2017fabricating, taylor2017cardiovascular}, space robotics~\cite{lamoureux2015dynamic, felton2014method} and educational tools~\cite{robogami, shi2025oribot}. In robotics, origami-inspired mechanisms have been integrated into a variety of designs capable of crawling~\cite{crawl2_onal, crawl3_koh, singh2024multi}, walking~\cite{baisch2014high, firouzeh2015robogami}, jumping~\cite{singh2024multi}, and grasping~\cite{chen2021soft, li2019vacuum}.

Despite these advances, the design of functional origami mechanisms remains largely intuitive and reliant on physical iteration. Modeling the relationship between crease patterns, material properties, and 3D kinematics requires solving for behavior over high dimensional, nonlinear parameter spaces~\cite{xue2025origami}, making analytical prediction computationally prohibitive or  mathematically complex. Achieving a desired motion or performance often requires repeated cycles of geometric tuning, fabrication, and testing, as small variations in crease geometry, fold angle, or sheet stiffness can lead to different mechanical behaviors~\cite{xi2023multi}, a process that could be consuming in both time and material. Recent studies have applied optimization techniques to improve the performance of origami inspired robotic systems. For example, Chen et al.~\cite{chen2023design} employed a genetic algorithm to optimize a foldable pneumatic actuator over multiple geometric parameters, achieving enhanced actuation performance. However, their approach relied on a detailed system specific mathematical model of the volume, pressure, and torque of their actuator. Cao et al.~\cite{cao2024design} proposed an inverse design framework for an origami gripper that adjusts crease pattern parameters to optimize for better grasping metrics, which proved very effective for task specific performance tuning. Yet, the framework relies on analytical modeling of a particular Yoshimura based geometry, making it less generalizable to other origami morphologies. Zhu et al.~\cite{10801958} proposed a bio-inspired trajectory optimization framework for a multi-locomotion origami robot, combining dynamic and kinematic modeling with graph based path planning to optimize gaits for crawling and swimming. Although this work demonstrates strong system level integration and effectively addresses its target locomotion tasks, its optimization is closely coupled to the robot’s specific morphology and motion modes. More recently, optimization methods have been extended to reconfigurable origami inspired manipulators, where physical reconfiguration parameters and control policies are co-optimized through reinforcement learning to enable adaptive behavior across tasks ~\cite{chen2025co}.
While these works showcase the potential of optimization in origami robotics, they depend on specialized forward models or analytical approximations that limit generalization to other morphologies or materials. Most existing methods require derivation of a custom kinematic model for each mechanism, a process that is both time consuming and not generalizable. While these models are capable of capturing local deformation or actuation behavior of a specific pattern with high physical fidelity, they typically cannot simulate how the structure interacts with external objects or dynamic environments without significant modification. This motivates the development of a more generalizable and accessible framework, one that allows designers to test and refine origami mechanisms directly within a realistic physics simulation environment, enabling rapid iteration and performance driven design exploration.

% For example, Chen et al.~\cite{10361525} used a genetic algorithm to optimize a foldable pneumatic actuator over multiple geometric parameters, while Cao et al.~\cite{cao2024design} proposed an inverse-design framework for an origami gripper capable of universal grasping. Zhu et al.~\cite{10801958} developed a bio-inspired trajectory optimization framework using kinematic modeling and graph-based path planning for an origami robot exhibiting both crawling and swimming locomotion. More recently, optimization methods have been extended to reconfigurable origami-inspired manipulators, where physical reconfiguration parameters and control policies are co-optimized through reinforcement learning to enable adaptive behavior across tasks~\cite{chen2025co}.

% While these approaches demonstrate effective design optimization, their underlying models are often tailored to specific morphologies and require detailed analytical understanding of structure kinematics and material behavior. This motivates the development of a more generalizable and accessible framework, one that allows designers to test and refine origami mechanisms directly within a realistic physics simulation environment, enabling rapid iteration and performance-driven design exploration.

To address this gap, we present a physics based origami simulation framework that integrates geometric design, dynamic modeling, and performance optimization within a single framework. Our system represents origami mechanisms as graphs of interconnected deformable panels, simulated using MuJoCo’s deformable body physics engine ~\cite{mujoco}. 
% Through a graphical user interface (GUI), users can intuitively define crease patterns, assign actuation and material constraints, and automatically generate physically consistent MuJoCo models. This setup allows for realistic dynamic simulations that capture both the folding mechanics and interactions with external objects or surfaces. Beyond visualization, our framework supports design optimization. By directly linking simulation results with algorithmic search. We demonstrate this through a case study on an origami catapult mechanism, where two geometric parameters are optimized using the Covariance Matrix Adaptation Evolution Strategy (CMA-ES). The resulting design achieves a significantly longer throwing distance, and hardware experiments confirm the simulation’s predictive fidelity. 
The contributions of this work are:
\begin{itemize}
    \item A physics based simulation framework for origami mechanisms that integrates geometric design, dynamic modeling, and optimization, enabling actuation driven simulations through MuJoCo’s deformable body system.
    \item A graph based representation and graphical user interface (GUI) that allow users to intuitively define crease patterns, assign material and actuation properties, and automatically generate physically consistent MuJoCo models from 2D schematic inputs.
    \item A case study demonstrating simulation driven optimization and hardware validation, where the framework is used to optimize an origami catapult mechanism via Matrix Adaptation Evolution Strategy (CMA-ES) \cite{hansen2023cma}, achieving improved throwing performance and consistent ranking between simulation and physical experiments.
\end{itemize}

An overview of the complete framework, spanning user design input, automatic model generation, simulation, and hardware validation, is illustrated in Fig.~\ref{fig:overall_fig}.

\section{Related Work}

 Recent advances in computationally efficient origami simulations focus on geometric and kinematic modeling of fold patterns, emphasizing mathematical tractability and visualization of crease patterns. For example, Demaine et al. established rigorous geometric models for understanding foldability and motion in origami-based systems ~\cite{demaine2007geometric}, while interactive tools like Origamizer~\cite{tachi2009origamizing,tachi2009simulation}, Rigid Folding Simulator~\cite{Tachi_RigidOrigami} and Freeform Origami ~\cite{Tachi_FreeformOrigami} enabled users to visualize folding trajectories that satisfy loop-closure constraints and fit any triangulated 3D surface using origami creases. 
 These tools allow for rapid design exploration and visualization of the origami mechanism, yet they are limited to quasi-static or kinematic folding and do not capture dynamic behaviors or actuation effects.

Methods have also been developed to incorporate mechanics and multiphysics-based modeling to simulate deformation, actuation, and nonlinear material responses. Ghassaei et al.~\cite{ghassaei2018fast} introduced a GPU-accelerated origami simulator that allows real time interaction on a web browser and visualization of material strain, while Sung et al.~\cite{sung2015computational} developed a computational design tool that enables automated creation of compliant mechanisms from origami patterns. Multiphysics simulators have also been developed to couple thermal, magnetic, or fluidic actuation to folding behavior~\cite{origami_sim_review}. While prior work has started to bridge the gap between geometric modeling and physical behavior, integrating origami mechanisms into general purpose robotic simulation environments remains an open challenge. Building on these efforts, we present a physics based origami simulation framework implemented in MuJoCo, a high fidelity, general purpose physics engine~\cite{mujoco}. Our approach enables dynamic simulation of origami mechanisms interacting with rigid bodies, actuators, and sensors within a unified system.

     \label{subsec:gui}
    \begin{figure}[h]
        \centering
\includegraphics[width=1\linewidth]{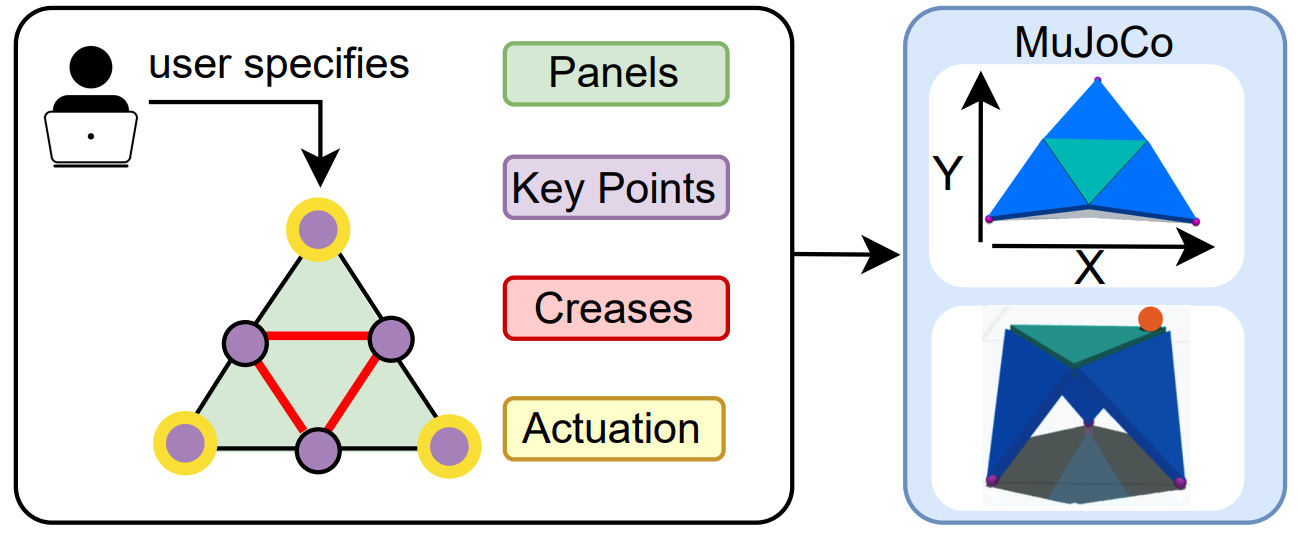}
        \caption{(Left) purple circles show the key points, yellow highlights key points with defined actuation, black lines are boundary edges, and red lines are creases. Our algorithm converts input from the GUI (left) into MuJoCo XML format to simulate the specified components (upper right), and by using the actuated key points at the very tip of each arm, we can raise and move the mechanism (lower right), dynamically interacting with a rigid sphere placed on the top panel of the mechanism.}
        \label{fig2}
    \end{figure}

\section{Methodology}
In this section, we first present a graphical user interface (GUI)~\cite{origami_framework_repo} developed to take intuitive definitions of an origami crease pattern and render the corresponding origami mechanisms in the physics simulator MuJoCo~\cite{mujoco}. We then showcase example mechanisms created using this framework, from their fold definitions to how they behave in the simulation.  Finally, we use select mechanisms to show 1) the simulation platform as a design tool that allows quick design iteration by sweeping through a range of design parameters, and 2) the optimization in simulation of an origami design using an evolution algorithm, Covariance Matrix Adaptation Evolution Strategy (CMA-ES), and a validation of the resulting design in hardware.

\subsection{Graphical User Interface for Design Rendering} 
    % To begin any design process, one must first decide how to define a desired structure and how to quickly realize it. 

    To facilitate the creation of origami mechanism designs, we developed a custom GUI that allows users to define the crease pattern of an origami mechanism abstracted as a two-dimensional graph structure. First, users define the vertices of the graph — key points used to specify boundaries and creases. Next, users add lines that connect the key points together and specify whether each line is a crease line, which means that the line will act as a hinge between the two adjacent panels, or, a boundary line, which defines the external boundary of a panel. To define the movement of panels around a crease, users assign properties to key points on the panels on each side of the crease line. The assignment defines the translational degrees of freedom on the x-, y-, or z-axis that each key point can have. Users can also designate key points as actuation points, specifying actuation along a chosen global axis (x, y, or z) in the simulation environment. Once the key points and lines form an enclosed geometry, the user defines panels by clicking within the closed shapes. All of the above user definitions are automatically converted into MuJoCo's XML-based MJCF format, enabling simulation of the corresponding deformable structure.

    An example input-output pair is shown in Fig.~\ref{fig2}. From the user definition provided to the GUI, we generate an origami blueprint of connected 2D panels. This allows the 2D sheet to exhibit the desired degrees of freedom determined by the specified folds and actuators. At this point, the origami mechanism is fully defined in the GUI and can be automatically exported as an XML file compatible with MuJoCo for rendering and simulation. When the user loads the exported XML file into MuJoCo, the mechanism appears as a flat sheet resting on the simulated ground plane under gravity. The environment includes a ground plane that constrains the motion of the structure through collision and friction. Because the actuation axes have already been defined in the GUI, users can now apply forces or control input directly to the actuated key points to fold the structure into its intended 3D configuration. The user can also reposition key points by modifying their coordinates in the XML file. In the example shown, this can be done by elevating the key points that form the central triangle. Alternatively, users can define the key points at the tips of the outer triangles to be actuation points, enabling the mechanism to fold and emerge into a 3D configuration from the 2D blueprint. This completes the basic transition from a 2D crease pattern to the 3D mechanism. Since, we render the mechanism in a physics environment, additional objects can be introduced into the environment; in this example, a sphere is placed on the top panel of the mechanism, dynamically responding to the movement of the structure rolling across its surface. 
    
    To clarify panel definition in detail, we choose the smallest enclosing cycle that includes at least one crease, then we sort the key points within each panel counterclockwise around their centroid to ensure consistent orientation prior to triangulation. Unlike convex hull algorithms, such as Graham Scan \cite{graham_scan_kong}, which discard interior points, our method retains all key points to preserve the structure of the original panel. This ensures that the panels are well formed, simple polygons suitable for further meshing. Preserving all key points within each panel is crucial for accurately capturing local geometric variations, which allows the resulting mesh to have some deformability in simulation. In addition, retaining this vertex-level structure lays the groundwork for modeling panels with varying material properties in future extensions of the framework. Each user selection results in the definition of one panel, and the process is repeated until all desired panels are specified.
    \begin{algorithm}[h]
    \caption{Panel Detection and Triangulation}
    \label{alg:face_mesh_generation}
    \begin{algorithmic}[1]
    \Require Graph $\mathcal{G}$ with vertices, edges, and fold labels
    \While{user clicks within a region}
    \State Detect minimal cycle enclosing the click position that contains at least one fold edge
    \State Sort the vertices of the detected cycle counterclockwise
    \State Record the panel defined by the sorted cycle
    \EndWhile
    \For{each recorded panel}
    \State Offset collinear vertices to avoid degenerate triangles
    \State Apply constrained Delaunay triangulation, preserving creases as fixed constraints
    \EndFor
    \Ensure Triangulated mesh structure for MuJoCo flexible-body simulation
    \end{algorithmic}
    \end{algorithm}

    Once the panels are constructed, we apply Delaunay triangulation~\cite{lee1980two} to each panel to generate a triangular mesh. Delaunay triangulation maximizes the minimum angle across the triangles, avoiding skinny triangles that can cause numerical instability during simulation. However, to triangulate a polygon, especially using Constrained Delaunay Triangulation, the polygon needs to be strictly simple, which requires that there are no overlapping edges and no three consecutive points perfectly collinear within one panel. The first requirement is automatically guaranteed as we do not store replicated edges. To ensure the second requirement, before triangulation, we apply a minor perturbation to collinear key points to prevent degenerate triangles during meshing. The triangulation results define each panel by referencing the key points' indices that make up the panel, rather than the modified coordinates of those key points, thus preserving the original mechanism while ensuring numerical stability during the triangulation process.
    The overall panel construction and meshing process is summarized in Algorithm~\ref{alg:face_mesh_generation}.
    
    % It is important to partition panels carefully around fold edges to accurately preserve folding behavior. For example, consider four vertices $\{v_1, v_2, v_3, v_4\}$ arranged in a square. If there is a fold edge defined between $v_2$ and $v_3$, then, instead of treating all four vertices as a single panel, we split the region into two triangular panels: $(v_1, v_2, v_3)$ and $(v_2, v_3, v_4)$. This ensures that the fold line is properly represented in the simulation. If the square were treated as a single panel, Delaunay triangulation would create a mesh that does not explicitly encode the fold, resulting in an unintended continuous soft body without folding behavior. This can be generalized to all polygons and both definitions can be incorporated into a larger mechanism. The user, however, is expected to define fold edge according to the design. 
    After generating the mesh for each face, the origami mechanism is instantiated in MuJoCo as a flexible deformable body, known as a ``flex". A flex element is represented as a collection of point masses (MuJoCo bodies) connected by elastic deformable elements (edges (1D), triangles (2D), or tetrahedra (3D)) that define its geometric and mechanical structure. Modeling origami as 2D flex sheets allows MuJoCo to discretize each panel into triangular elements that support dynamic interaction and contact resolution. While recent versions of MuJoCo (3.5) provide a Saint Venant–Kirchhoff elasticity model for 2D flex sheets, at the time our framework was developed (MuJoCo 3.2) there was more limited functionality in this regard. Consequently, we did not explicitly calibrate membrane material parameters (e.g., Young’s modulus or Poisson’s ratio) to match the physical material properties of the fabricated prototypes. Instead, we leverage the flex sheets to enable geometric coupling and dynamics. In our framework, each panel of the origami mechanism is modeled as an individual flex sheets, with the user defined key points serving as the point mass bodies with triangulation completed by Algorithm~\ref{alg:face_mesh_generation}. Because flex sheets act directly on the underlying bodies in the model, panels that share key points implicitly couple through those shared bodies, enabling the simulation of dynamic interactions across the mechanism.

    % After generating the mesh for each face, the origami mechanism is instantiated in MuJoCo as a flexible deformable body, known as a ``flex". MuJoCo versions 3.0 and above introduce a physics-based framework for modeling deformable bodies through a flex system.  A flex object is represented as a collection of point masses (MuJoCo bodies) connected by elastic deformable elements (edges (1D), triangles (2D), or tetrahedra (3D)) that define its geometric and mechanical structure. Modeling origami as 2D flex sheets enables the MuJoCo platform to discretize surfaces into triangular elements,  that are treated as constant-stress finite element governed by the Saint Venant–Kirchhoff hyperelastic constitutive law~\cite{6386109}. In our framework, each panel of the origami mechanism is modeled as an individual flex object, with the user-defined key points serving as the point-mass bodies, and the triangulation is completed by the algorithm described above. Because flexes act directly on the underlying bodies in the model, panels that share key points implicitly couple through those shared bodies. The motion of these shared points reflects the combined elastic and contact forces from all connected panels, enabling the simulation of dynamic interactions across the origami mechanism. 

    \subsection{Example Mechanisms \& Key features}

    \begin{figure}[H]
        \centering
        \includegraphics[width=0.95\linewidth]{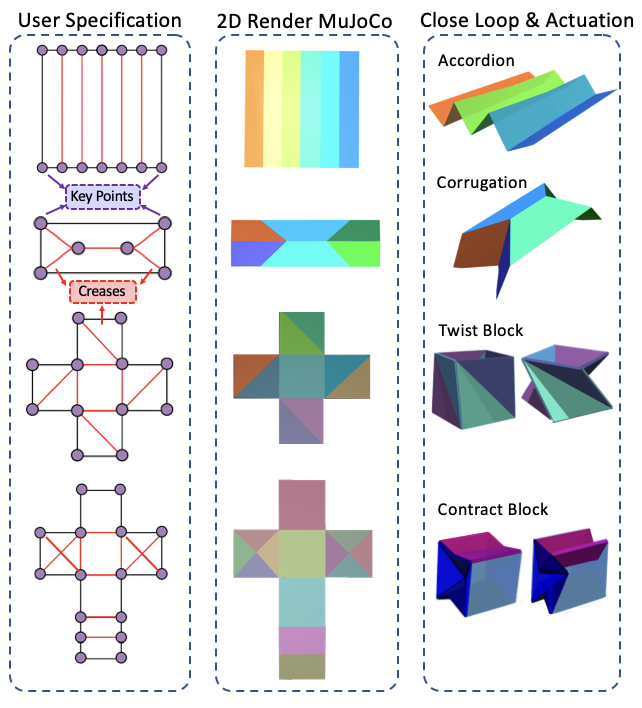}
        \caption{Examples of standard origami fold patterns processed through our framework. The left column shows user-defined specifications in the graphical interface, where key points (purple) and crease lines (red) define the geometric relationships between panels (actuations omitted for clarity). The middle column shows the corresponding 2D renderings of the generated MuJoCo flex sheets, and the right column shows the resulting 3D closed and actuated mechanisms.  From top to bottom: (1) parallel strip folds forming an accordion-like structure; (2) a corrugation pattern with V-shaped valley folds supporting a central mountain ridge; (3) a modular origami actuator block achieving rotational motion through coordinated folding; and (4) a modular origami actuator block achieving horizontal contraction.}
        \label{fig:def2fold}
    \end{figure}
    
    To evaluate the modeling capabilities of our framework, we begin by simulating standard origami fold patterns and mechanisms from folding schematics, through our GUI, to simulated mechanisms rendered in simulation. We refer to mechanisms that contain panels connected in a cyclic manner (e.g. a box) as a closed-loop mechanisms. In Fig.~\ref{fig:def2fold}, we present this process for both open and closed-loop mechanisms. While both open and closed-loop mechanisms can be defined using our GUI and rendered in simulation, we note that the two closed-loop mechanisms (bottom two in Fig.~\ref{fig:def2fold}) would require the user to manually merge desired key points where panels join in the XML document to ensure the mechanism closes as intended. These examples illustrate the complete workflow of our framework: from defining the crease pattern to generating the resulting three-dimensional mechanism.
    
\begin{figure}[H]
        \centering
        \includegraphics[width=1\linewidth]{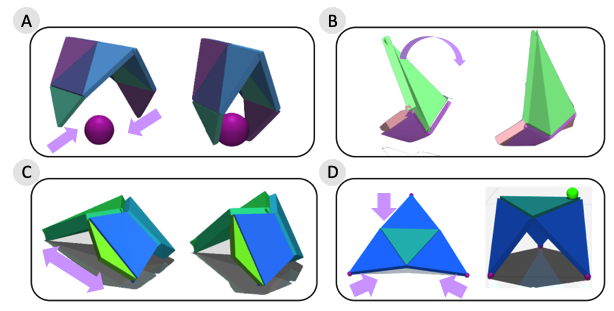}
        \caption{Examples of origami mechanisms interacting with their environment. (A) An origami gripper that can push, grasp, and release an object. (B) An origami catapult where the top panel rotates upward under lateral actuation. (C) A walker that propels itself forward on the ground plane through cyclic contraction and release. (D) A triangular, legged origami mechanism capable of balancing a rolling sphere by modulating actuation across its legs. The purple arrows indicate the direction of motion of each structure.}
        \label{fig:interactive_structs}
    \end{figure}

    Our simulation framework also captures interactions between origami mechanisms and their surrounding environment, such as rigid objects or the ground plane. As shown in Fig.~\ref{fig:interactive_structs}, the framework can model various behaviors including grasping, throwing, locomotion, and balancing. These examples highlight how our framework enables dynamic interaction between origami mechanisms and external objects.

    \subsection{Using the Framework for Optimization}
    \begin{figure}[ht]
        \centering
        \includegraphics[width=1\linewidth]{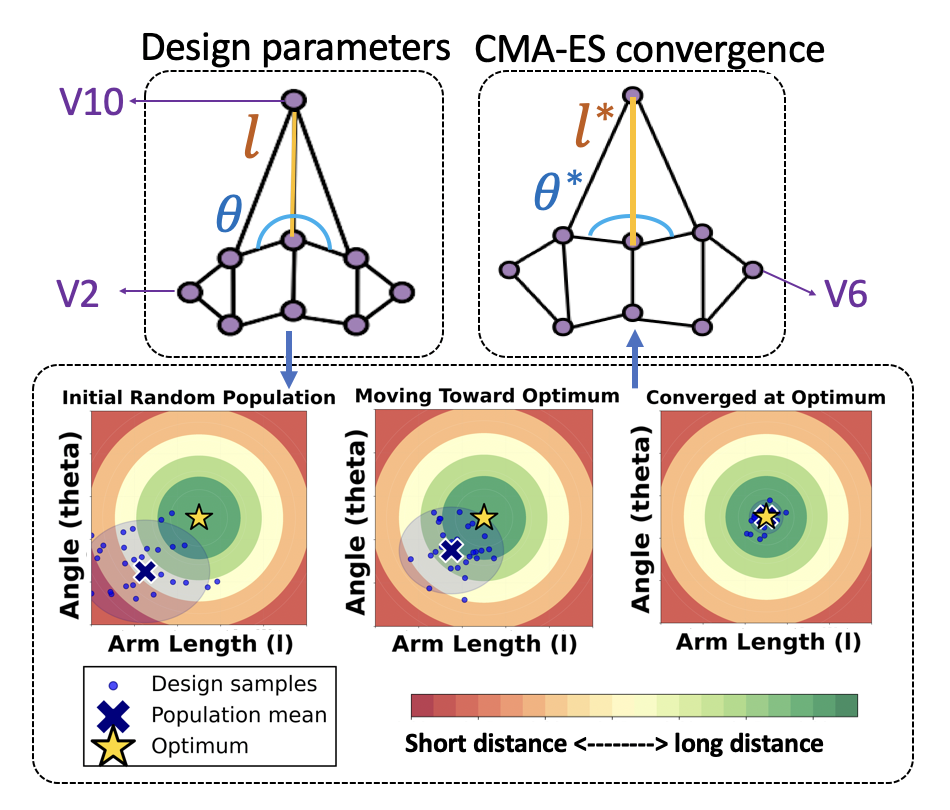}
        \caption{Optimization of the origami catapult mechanism over two design parameters: (1) the sector angle of the mountain folds ($\theta$) and (2) the length of the throwing arm ($l$). The top row shows the initial and optimized configurations of the mechanism. The bottom row illustrates the CMA-ES optimization process, where the population of candidate designs progressively converges toward the region corresponding to maximal throwing distance.}
        \label{fig:5}
    \end{figure}
    Beyond serving as a simulation and visualization tool, our framework enables design optimization by allowing structural parameters to be systematically altered and evaluated in physics-based simulation. To demonstrate this capability, we simulate an origami mechanism, catapult, where its dynamic behaviors arise from force propagation and mechanical energy transfer (mechanism B of Fig.~\ref{fig:interactive_structs}). In the origami catapult mechanism, lateral actuation is converted into vertical launch force. 
    
    To demonstrate rapid optimization over potential designs, we apply the Covariance Matrix Adaptation Evolution Strategy (CMA-ES), a stochastic derivative-free optimization algorithm well suited for non-linear, non-convex search spaces. CMA-ES maintains a population of design candidates, samples new candidates based on a multivariate normal distribution, and iteratively updates the distribution parameters toward regions of higher performance. An illustration of this design optimization process is shown in Fig.~\ref{fig:5}. We conduct our optimization in Python using the CMA-ES library with the step size set to $\sigma = 0.025$.
\section{Simulation Results}

We present the design and optimization results of our framework on the catapult mechanism. Two key geometric parameters were varied to generate new designs:
\begin{itemize}
    \item Fold angle $\theta$ of the mountain fold, which is controlled by 2 key points (vertices 2 and 6 in Fig.~\ref{fig:5}). We vary this parameter by systematically changing the position of vertices 2 and 6, allowing us to search over the range of $\theta \in [100^{\circ}, 226^{\circ}]$
    \item Arm length $l$ is controlled by one key point (vertex 10 in Fig.~\ref{fig:5}). We search over the range $l \in [8, 18]$ centimeters.
\end{itemize}

We first design, define, and render the origami catapult structure using the GUI and workflow described in the previous section. We then apply the CMA-ES \cite{hansen2019pycma} optimization to the simulated catapult mechanism introduced in Section~III. The objective of the simulation is to maximize the horizontal throwing distance of a spherical payload. To ensure comparability between simulation and hardware experiments, key geometric and actuation parameters were matched across both domains. In addition, the spherical projectile's mass ($1~\mathrm{g}$) and radius ($0.01~\mathrm{m}$) were also identical. In hardware, two Dynamixel motors actuated the mechanism with a synchronized outward rotation of $75^{\circ}$, achieving an average angular velocity of approximately $2.09~\mathrm{rad/s}$ while the simulation actuation command was tuned to produce a comparable angular velocity around $2.08~\mathrm{rad/s}$ at the actuated key points. Gravity was modeled identically in simulation. Contact interactions between the sphere and the catapult arm were resolved through MuJoCo’s contact solver, while hardware interactions arose from physical contact between PETG surfaces.

 To better analyze the optimization results, we also perform a parameter sweep. For each design, we vary the length of the throwing arm of the catapult $l$ as well as the angle of the fold $\theta$. We search over the same ranges: $\theta \in [100^{\circ} , 226^{\circ}]$ degrees and arm length $l \in [8, 18]$ centimeters. We iterate over 2880 combinations of parameters $\theta$ and $l$. In both the brute force parameter sweep and the CMA-ES optimization, other than the key points involved in altering these parameters (two vertices controlling the angle and one vertex controlling the arm length), all key points are kept in the same position with the same properties to ensure consistency. 

 \begin{figure}[h]
     \centering   \includegraphics[width=0.98\linewidth]{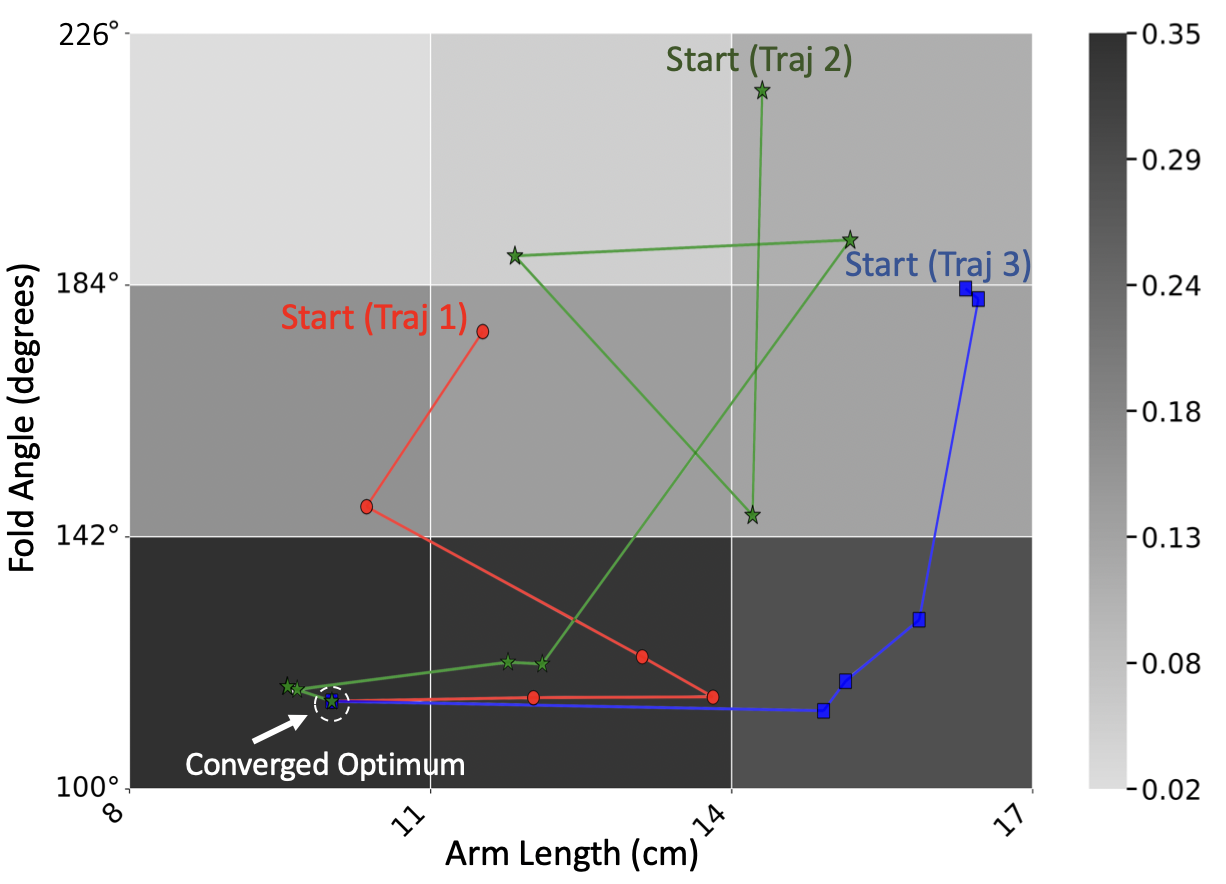}
     \caption{Heatmap of simulated throwing distance across the design space defined by fold angle ($\theta$) and arm length ($l$) found with parameter sweep, overlaid with CMA-ES optimization trajectories. Each bin of the heat map represents a subrange of the parameters with its color representing the average throwing distance of simulated designs within that bin. Colored paths trace the progress of three independent CMA-ES runs over 200 generations, generation 0, 1, 2, 4, 8, 16, 32, 64, 128, and 200. All trajectories converge toward the same high-performance region and ultimately to the same optimal design. }
     \label{fig:heatmap}
 \end{figure} 

The parameter sweep highlights a clear performance group corresponding to sharper fold angles ($100^{\circ}$ to $142^{\circ}$) and moderate arm lengths ($8 ~\mathrm{cm}$ to $11~\mathrm{cm}$) with the best design ($\theta = 119.3^{\circ}$ and $l = 9.6~\mathrm{cm}$) achieving a distance of $46.4~\mathrm{cm}$, suggesting that insufficiently sharp fold curvature or overly long arms reduce the energy transfer efficiency from lateral actuation to vertical launch. However, because each bin represents the average performance of all designs that fall within its discrete interval, the true optimum may not be captured precisely. Achieving a finer resolution would require sampling a substantially larger number of parameter combinations, rapidly increasing the computational cost of an exhaustive search. To evaluate sample efficient design optimization, we ran CMA-ES from randomized initial conditions over the same parameter space. We showcase four such optimization trajectories in Fig.~\ref{fig:heatmap} as a colored path overlaid on the heatmap of sweep results. All trajectories converge toward the same region identified by the brute-force search, validating the consistency and robustness of the simulation based optimization pipeline. The best-performing configuration identified by CMA-ES corresponds to a fold angle ${\theta}^{*} \approx 115.5^{\circ}$ and arm length $l^* \approx 10.2~\mathrm{cm}$, achieving a simulated average throwing distance of  $47.2~\mathrm{cm}$. %This design was later reproduced in hardware for validation experiments Fig.~\ref{fig:hardware_designs}.

\section{Hardware Validation Experiments and Results}

\subsection{Motivation}

The motivation of the hardware experiments is to validate the predictive accuracy of our simulation framework by comparing the performance of the optimized mechanism identified in simulation using the CMA-ES optimization against two sub-optimal designs predicted to perform worse in simulation. While we expect quantitative differences due to sim-to-real discrepancies arising from unmodeled material properties and imperfect representation of crease compliance, the relative performance ordering should remain consistent, as the geometry of each design dictates its throwing behavior, which should be captured by simulation. 

\begin{figure}[H]
\centering
\includegraphics[width=0.90\linewidth]{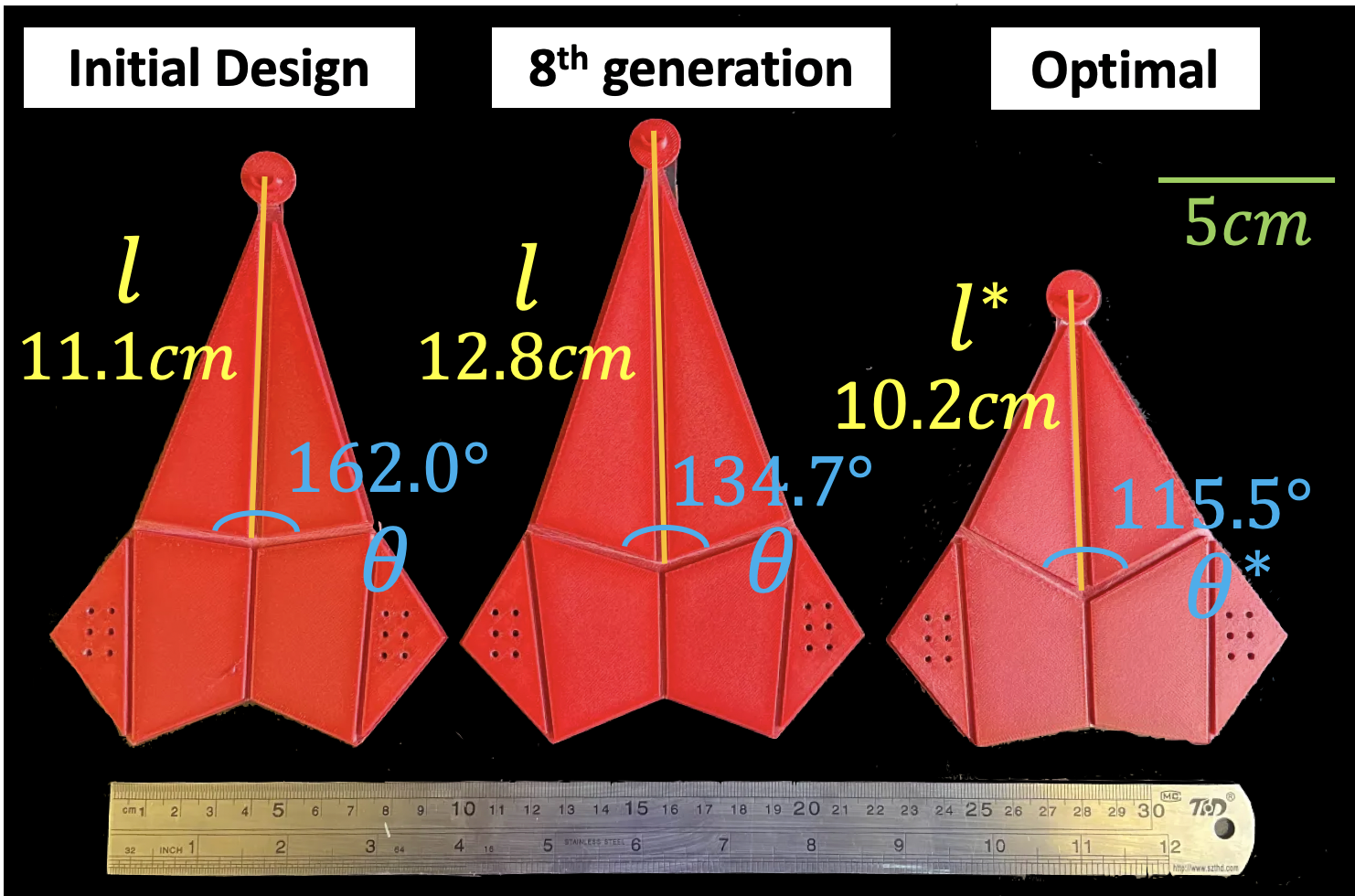}
\caption{Three catapult configurations selected for hardware validation. The optimized design corresponds to the best performing parameters found in simulation. We also included the initial design that the optimization process started with, and the best performing design we found at the $8^{th}$ generation of the optimization trajectory.}
\label{fig:all_designs}
\vspace{-1em}
\end{figure}

We sampled a single CMA-ES optimization trajectory and recorded three representative designs: the randomly initialized starting design, the best-performing design from the $8^{th}$ generation, and the final optimal design (Fig.~\ref{fig:all_designs}). The optimization results demonstrate progressive improvement from the initial to the $8^{th}$ generation to the final optimized design, and we expect this performance ordering to be consistent in the real-world validation experiments. The $8^{th}$ generation was chosen because as Fig.~\ref{fig:heatmap} shows, there is greater variability in the design between generation $4$ and $16$ where the designs is optimized but has not yet converged to be close to the optimum. 

\subsection{Experimental Design \& Setup}
\begin{figure}[H]
\centering
\includegraphics[width=0.7\linewidth]{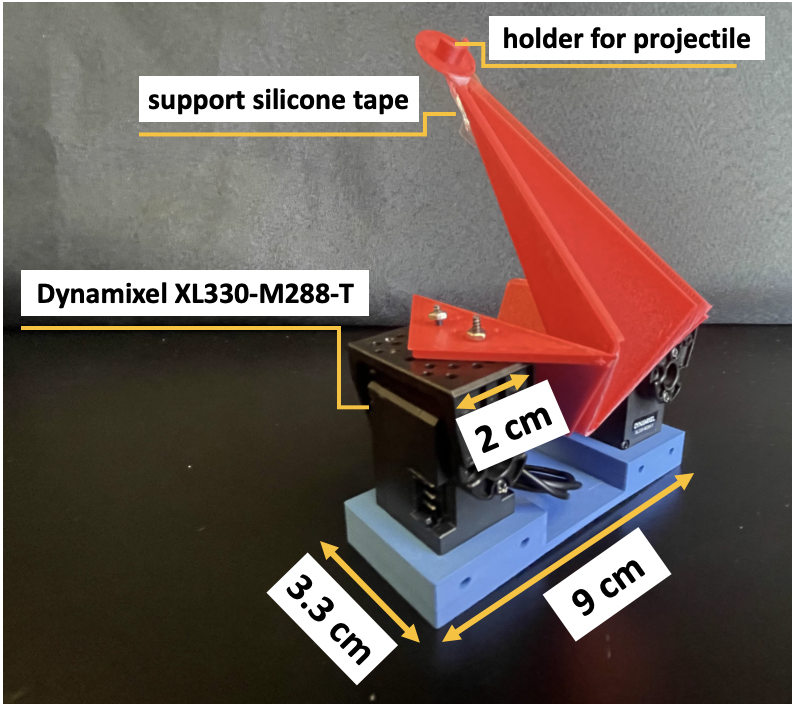}
\caption{Hardware setup. Two Dynamixel motors rigidly mounted to a $90 ~\mathrm{mm}$ base actuate the origami catapult synchronously. Each motor holds a $2~\mathrm{cm}$ mounting basket to which the origami mechanism is fastened using $2~\mathrm{mm}$ M2 screws. Positional control commands rotate both motors $75^{\circ}$ outward, replicating the actuation profile used in simulation.}
\label{fig:hardware_designs}
% \vspace{-1em}
\end{figure}

The three catapult prototypes, corresponding to the initial, $8^{th}$ generation, and final optimized design (Fig.~\ref{fig:all_designs}), were 3D-printed using the same fold geometry and spacing as in simulation. We printed the designs using PETG on a Bambu H2D 3D-printer. The faces are made with thickness of $2~\mathrm{mm}$ to allow for some rigidity that makes the mechanism more robust and less likely to be damaged by actuation, and the folds are made to be $0.2~\mathrm{mm}$ to allow flexible folding behavior.

Each mechanism was mounted, as shown in Fig.~\ref{fig:hardware_designs} to a rigid PLA base and actuated using two Dynamixel motors controlled in position mode. The motors were tightly fitted to a base mount of width $9~\mathrm{cm}$ and rotated synchronously by $75^{\circ}$ outward to generate the catapulting motion.

% IDK if we should mention but  (simulation achieves around $2.078$ rad/s). The thing is, the motors are also inconsistent sometimes it goes higher or lower but it is always close to 2 rad/s) 

\subsection{Hardware Experiment Results}
Each design was tested for three trials under identical actuation parameters. The mean and standard deviation of projectile travel distance are summarized in Table \ref{tab:a}. As predicted by simulation, the optimized design achieved the greatest throwing distance, followed by the two sub-optimal variants.

\begin{table}[h]
    \centering
    \caption{Mean Distance Projectile Traveled in Hardware Validation Experiment}
    \begin{tabular}{lccc}
        \toprule
        \textbf{Results} & \textbf{Initial Design} & \textbf{$8^{\text{th}}$ Gen} & \textbf{Opt Design} \\
        \midrule
        Mean Dist (cm) & 30.80 & 52.90 & 57.00 \\
        Std Dev (cm) & 4.29 & 0.52  & 1.81 \\
        \bottomrule
    \end{tabular}
    \label{tab:a}
    \vspace{-1em}
\end{table}

\begin{figure}[H]
\centering
\includegraphics[width=1\linewidth]{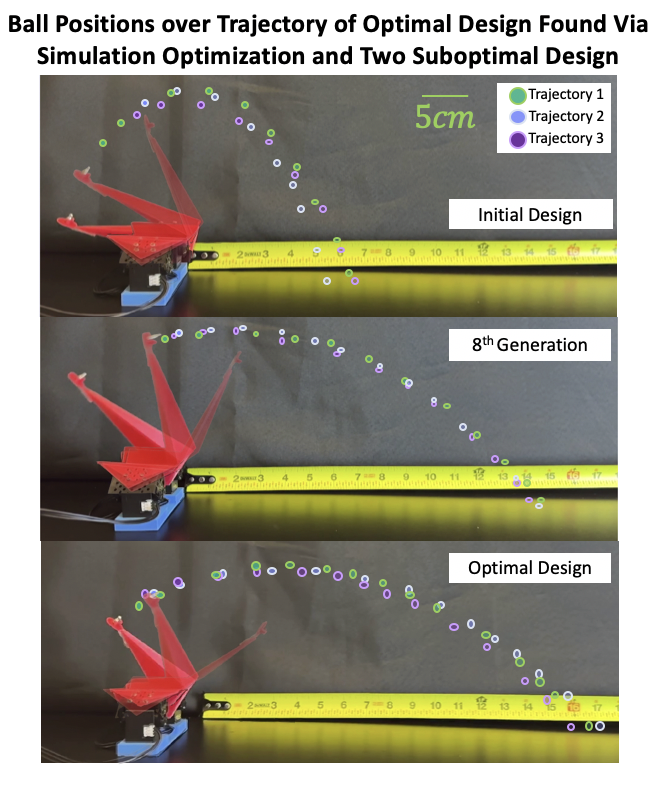}
\caption{Projectile trajectories for three catapult designs recorded using frame tracking software, where we mark the path of the projectile using key frames. The optimized configuration achieves the longest throws, consistent with simulation predictions. The tape measure in the background has markings in inches. }
\label{fig:traj_comp}
\vspace{-1em}
\end{figure}

The projectile trajectories recorded by a camera using slow motion at $240$ frames per second are shown in Fig.~\ref{fig:traj_comp}, illustrating consistent ordering of performance between designs. The waypoints were marked using the trajectory of the projectile captured in the videos. 
\vspace{-1em}

To further assess simulation fidelity, Table \ref{tab:distance_comparison} compares the mean throwing distances obtained in simulation and hardware, along with percentage differences. Although absolute values differ due to unmodeled factors such as air resistance and material imperfections, the relative ranking of the designs remains consistent across both domains, demonstrating the validity of our simulation framework for comparative design evaluation.

\begin{table}[h]
\centering
% \caption{Comparison of Throwing Distance Between Simulation and Hardware with Sim to Real Difference and Performance Improvement}
\caption{Simulation to real performance difference}
\begin{tabular}{lccc}
\toprule
\textbf{Design} & \textbf{Sim (cm)} & \textbf{Real (cm)} & \textbf{Diff (\%)} \\
\midrule
Initial Design & 10.84 & 30.80 & 64.8\%  \\
$8^{\text{th}}$ Gen & 42.77 & 52.90 & 19.1\%\\
Optimal Design & 47.17 & 57.00 & 17.2\%\\
\bottomrule
\end{tabular}
\label{tab:distance_comparison}
\vspace{-1em}
\end{table}

The experimental results corroborate the simulation predictions: the design identified as optimal in simulation maintained superior performance in physical testing, while the sub-optimal configurations produced consistently shorter launch distances. Although quantitative discrepancies exist between simulated and real measurements, likely due to simplifications in material modeling and actuation dynamics not yet fully supported in the simulation platform, the framework successfully captured the relative performance trends based on the geometry of the design.

\section{Conclusion}
In this work, we developed a simulation framework that enables the design and dynamic modeling of origami mechanisms directly within a physics based environment. Our system converts user-defined GUI designs into deformable MuJoCo models, automatically generating meshes, fold constraints, and actuation parameters defined by the user for dynamic simulation. We demonstrated that the framework can capture folding behaviors and interactions with the environment, allowing for both visualization and functional analysis of origami based robots. To illustrate its utility for design exploration, we applied the framework to optimize an origami catapult mechanism with the objective of maximizing projectile distance. The initial design, mid-optimization design, and resulting optimized design were then tested in hardware models, and the experimental results preserved the same relative performance ranking as observed in simulation. 

Although some quantitative discrepancies exist between simulated and real distances, this sim-to-real gap is a common challenge. In our case, we attribute these differences primarily to unmodeled factors such as material properties (e.g., Young’s modulus, crease stiffness, and local spring forces), manufacturing imperfections, and the abstraction of actuation in simulation, limitations that stem partly from the current parameterization supported by the MuJoCo deformable model we used. While the flex model captures in-plane deformation and dynamic coupling between connected panels, it does not allow direct specification of crease stiffness or bending energy in the current version of the software. Consequently, folds behave as compliant connections mediated through shared key points rather than as controllable hinges, which limits the software's ability to model crease integrity or programmable folding stiffness. Although there exist more origami specific simulators that enable detailed modeling of creases, materials, and folding mechanics, our framework prioritizes dynamic interaction, allowing origami mechanisms to be simulated alongside rigid bodies, actuators, and sensors within a unified physics based environment.
% This facilitates design-space exploration and reinforcement-learning-based control in a physically grounded environment.

% Our framework prioritizes dynamic interaction—allowing origami mechanisms to be simulated alongside rigid bodies, actuators, and sensors within MuJoCo—thus enabling design-space exploration and reinforcement-learning-based control in a physically grounded setting. 

The consistent relative ranking between simulated and hardware designs validates the framework’s predictive accuracy for comparative design evaluation. As an extension of this work, we aim to improve the physical fidelity of our models through better characterization of material and fold properties, as well as to explore integration with learning based or model driven optimization methods. Ultimately, we envision this framework as a general purpose platform for rapid, simulation driven design and analysis of origami inspired robotic mechanisms, encompassing both compliant and rigid interactive structures, paving the way for studying origami mechanisms in complex, task oriented contexts.

% \begin{itemize}
%     \item We developed a framework that takes...
%     \item we have simulated structures that... 
%     \item We then performed optimization on one structure with an objective to... 
%     \item In the hardware validation we found consistent ranking of performance however, we see some descrepency of sim and real results. Sim to real gap is a problem in robotics in general, we think the reason of the gap here is.. 
%     \item As an extension of this work, we want to look into ways to better...
% \end{itemize}


%%%%%%%%%%%%%%%%%%%%%%%%%%%%%%%%%%%%%%%%%%%%%%%%%%%%%%%%%%%%%%%%%%%%%%%%%%%%%%%%

%%%%%%%%%%%%%%%%%%%%%%%%%%%%%%%%%%%%%%%%%%%%%%%%%%%%%%%%%%%%%%%%%%%%%%%%%%%%%%%%

%%%%%%%%%%%%%%%%%%%%%%%%%%%%%%%%%%%%%%%%%%%%%%%%%%%%%%%%%%%%%%%%%%%%%%%%%%%%%%%%

%%%%%%%%%%%%%%%%%%%%%%%%%%%%%%%%%%%%%%%%%%%%%%%%%%%%%%%%%%%%%%%%%%%%%%%%%%%%%%%%

\renewcommand*{\bibfont}{\small}
% \newpage
\printbibliography

@misc{hansen2023cma,
title={The CMA Evolution Strategy: A Tutorial},
author={Nikolaus Hansen},
year={2023},
eprint={1604.00772},
archivePrefix={arXiv},
primaryClass={cs.LG}
}

@inproceedings{sung2015computational,
  author    = {Sung, Cynthia R. and Kuribayashi-Shigetomi, Kaori and An, Byoungkwon and Boyvat, Mustafa and Rus, Daniela},
  title     = {A Computational Design Tool for Compliant Mechanisms Based on Origami},
  booktitle = {Proceedings of the ASME 2015 International Design Engineering Technical Conferences and Computers and Information in Engineering Conference},
  year      = {2015},
  publisher = {ASME},
  pages     = {V05AT08A031}
}

@book{demaine2007geometric,
  title={Geometric folding algorithms: linkages, origami, polyhedra},
  author={Demaine, Erik D and O'Rourke, Joseph},
  year={2007},
  publisher={Cambridge university press}
}

@article{ghassaei2018fast,
  title={Fast, interactive origami simulation using GPU computation},
  author={Ghassaei, Amanda and Demaine, Erik D and Gershenfeld, Neil},
  journal={Origami},
  volume={7},
  pages={1151--1166},
  year={2018}
}

@article{rus2018design,
  title={Design, fabrication and control of origami robots},
  author={Rus, Daniela and Tolley, Michael T},
  journal={Nature Reviews Materials},
  volume={3},
  number={6},
  pages={101--112},
  year={2018},
  publisher={Nature Publishing Group UK London}
}

@INPROCEEDINGS{mujoco,
  author={Todorov, Emanuel and Erez, Tom and Tassa, Yuval},
  booktitle={2012 IEEE/RSJ International Conference on Intelligent Robots and Systems}, 
  title={MuJoCo: A physics engine for model-based control}, 
  year={2012},
  volume={},
  number={},
  pages={5026-5033},
  doi={10.1109/IROS.2012.6386109}}

@article{origami_sim_review,
author = {Zhu, Yi and Schenk, Mark and Filipov, Evgueni},
year = {2022},
month = {07},
pages = {},
title = {A Review On Origami Simulations: From Kinematics, to Mechanics, Towards Multi-Physics},
volume = {74},
journal = {Applied Mechanics Reviews}
}

@article{xi2023multi,
  title={Multi-stability of the extensible origami structures},
  author={Xi, Kaili and Chai, Sibo and Ma, Jiayao and Chen, Yan},
  journal={Advanced Science},
  volume={10},
  number={29},
  pages={2303454},
  year={2023},
  publisher={Wiley Online Library}
}

@article{graham_scan_kong,
title = {The Graham scan triangulates simple polygons},
journal = {Pattern Recognition Letters},
volume = {11},
number = {11},
pages = {713-716},
year = {1990},
issn = {0167-8655},
doi = {https://doi.org/10.1016/0167-8655(90)90089-K},
url = {https://www.sciencedirect.com/science/article/pii/016786559090089K},
author = {Xianshu Kong and Hazel Everett and Godfried Toussaint}
}

@article{robogami,
author = {Adriana Schulz and Cynthia Sung and Andrew Spielberg and Wei Zhao and Robin Cheng and Eitan Grinspun and Daniela Rus and Wojciech Matusik},
title ={Interactive robogami: An end-to-end system for design of robots with ground locomotion},

journal = {The International Journal of Robotics Research},
volume = {36},
number = {10},
pages = {1131-1147},
year = {2017},
}

@article{lamoureux2015dynamic,
  title={Dynamic kirigami structures for integrated solar tracking},
  author={Lamoureux, Aaron and Lee, Kyusang and Shlian, Matthew and Forrest, Stephen R and Shtein, Max},
  journal={Nature communications},
  volume={6},
  number={1},
  pages={8092},
  year={2015},
  publisher={Nature Publishing Group UK London}
}

@article{felton2014method,
  title={A method for building self-folding machines},
  author={Felton, Samuel and Tolley, Michael and Demaine, Erik and Rus, Daniela and Wood, Robert},
  journal={Science},
  volume={345},
  number={6197},
  pages={644--646},
  year={2014},
  publisher={American Association for the Advancement of Science}
}

@article{gultepe2012biopsy,
  title={Biopsy with thermally-responsive untethered microtools},
  author={Gultepe, Evin and Randhawa, Jatinder S and Kadam, Sachin and Yamanaka, Sumitaka and Selaru, Florin M and Shin, Eun J and Kalloo, Anthony N and Gracias, David H},
  journal={Advanced Materials (Deerfield Beach, Fla.)},
  volume={25},
  number={4},
  pages={10--1002},
  year={2012}
}

@article{crawl2_onal,
  title={An origami-inspired approach to worm robots},
  author={Onal, Cagdas D and Wood, Robert J and Rus, Daniela},
  journal={IEEE/ASME transactions on mechatronics},
  volume={18},
  number={2},
  pages={430--438},
  year={2012},
  publisher={IEEE}
}

@article{crawl3_koh,
  title={Omega-shaped inchworm-inspired crawling robot with large-index-and-pitch (LIP) SMA spring actuators},
  author={Koh, Je-Sung and Cho, Kyu-Jin},
  journal={IEEE/ASME Transactions On Mechatronics},
  volume={18},
  number={2},
  pages={419--429},
  year={2012},
  publisher={IEEE}
}

@article{baisch2014high,
  title={High speed locomotion for a quadrupedal microrobot},
  author={Baisch, Andrew T and Ozcan, Onur and Goldberg, Benjamin and Ithier, Daniel and Wood, Robert J},
  journal={The International Journal of Robotics Research},
  volume={33},
  number={8},
  pages={1063--1082},
  year={2014},
  publisher={SAGE Publications Sage UK: London, England}
}

@article{firouzeh2015robogami,
  title={Robogami: A fully integrated low-profile robotic origami},
  author={Firouzeh, Amir and Paik, Jamie},
  journal={Journal of Mechanisms and Robotics},
  volume={7},
  number={2},
  pages={021009},
  year={2015},
  publisher={American Society of Mechanical Engineers}
}

@inproceedings{singh2024multi,
  title={Multi-modal jumping and crawling in an autonomous, springtail-inspired microrobot},
  author={Singh, Shashwat and Temel, Zeynep and Pierre, Ryan St},
  booktitle={2024 IEEE International Conference on Robotics and Automation (ICRA)},
  pages={5999--6005},
  year={2024},
  organization={IEEE}
}

@article{xue2025origami,
  title={Origami Robots: Design, Actuation, and 3D Printing Methods},
  author={Xue, Wenbo and Jian, Bingcong and Jin, Liuchao and Wang, Rong and Ge, Qi},
  journal={Advanced Materials Technologies},
  pages={e00278},
  year={2025},
  publisher={Wiley Online Library}
}

@article{chen2021soft,
  title={Soft origami gripper with variable effective length},
  author={Chen, Bohan and Shao, Zhuyin and Xie, Zhexin and Liu, Jiaqi and Pan, Fei and He, Liwen and Zhang, Li and Zhang, Yanming and Ling, Xuechen and Peng, Fujun and others},
  journal={Advanced Intelligent Systems},
  volume={3},
  number={10},
  pages={2000251},
  year={2021},
  publisher={Wiley Online Library}
}

@inproceedings{li2019vacuum,
  title={A vacuum-driven origami “magic-ball” soft gripper},
  author={Li, Shuguang and Stampfli, John J and Xu, Helen J and Malkin, Elian and Diaz, Evelin Villegas and Rus, Daniela and Wood, Robert J},
  booktitle={2019 International Conference on Robotics and Automation (ICRA)},
  pages={7401--7408},
  year={2019},
  organization={IEEE}
}

@article{chen2023design,
  title={Design and optimization of an origami-inspired foldable pneumatic actuator},
  author={Chen, Huaiyuan and Ma, Yiyuan and Chen, Weidong},
  journal={IEEE Robotics and Automation Letters},
  volume={9},
  number={2},
  pages={1278--1285},
  year={2023},
  publisher={IEEE}
}

@article{cao2024design,
  title={Design and optimization of an origami gripper for versatile grasping and manipulation},
  author={Cao, Hanwen and Zhou, Jianshu and Chen, Kai and He, Qiguang and Dou, Qi and Liu, Yun-Hui},
  journal={Advanced Intelligent Systems},
  volume={6},
  number={12},
  pages={2400271},
  year={2024},
  publisher={Wiley Online Library}
}

@INPROCEEDINGS{10801958,
  author={Zhu, Keqi and Guo, Haotian and Yu, Wei and Nigatu, Hassen and Li, Tong and Dong, Ruihong and Dong, Huixu},
  booktitle={2024 IEEE/RSJ International Conference on Intelligent Robots and Systems (IROS)}, 
  title={Theoretical Modeling and Bio-inspired Trajectory Optimization of A Multiple-locomotion Origami Robot}, 
  year={2024},
  volume={},
  number={},
  pages={11355-11361}}

@article{chen2025co,
  title={Co-optimizing Physical Reconfiguration Parameters and Controllers for an Origami-inspired Reconfigurable Manipulator},
  author={Chen, Zhe and Chen, Li and Zhang, Hao and Zhao, Jianguo},
  journal={arXiv preprint arXiv:2504.10474},
  year={2025}
}

@article{tachi2009simulation,
  title={Simulation of rigid origami},
  author={Tachi, Tomohiro},
  journal={Origami},
  volume={4},
  number={08},
  pages={175--187},
  year={2009},
  publisher={Kerswell Books}
}

@article{tachi2009origamizing,
  title={Origamizing polyhedral surfaces},
  author={Tachi, Tomohiro},
  journal={IEEE transactions on visualization and computer graphics},
  volume={16},
  number={2},
  pages={298--311},
  year={2009},
  publisher={IEEE}
}

@misc{Tachi_FreeformOrigami,
  author       = {Tomohiro Tachi},
  title        = {Freeform Origami},
  howpublished = {\url{http://www.tsg.ne.jp/TT/software/}},
  note         = {Accessed: October 2025}
}

@misc{Tachi_RigidOrigami,
  author       = {Tomohiro Tachi},
  title        = {Rigid Origami Simulator},
  howpublished = {\url{http://www.tsg.ne.jp/TT/software/}},
  note         = {Accessed: October 2025}
}

@article{turner2016review,
  title={A review of origami applications in mechanical engineering},
  author={Turner, Nicholas and Goodwine, Bill and Sen, Mihir},
  journal={Proceedings of the Institution of Mechanical Engineers, Part C: Journal of Mechanical Engineering Science},
  volume={230},
  number={14},
  pages={2345--2362},
  year={2016},
  publisher={SAGE Publications Sage UK: London, England}
}

@article{lee1980two,
  title={Two algorithms for constructing a Delaunay triangulation},
  author={Lee, Der-Tsai and Schachter, Bruce J},
  journal={International Journal of Computer \& Information Sciences},
  volume={9},
  number={3},
  pages={219--242},
  year={1980},
  publisher={Springer}
}

@article{johnson2017fabricating,
  title={Fabricating biomedical origami: a state-of-the-art review},
  author={Johnson, Meredith and Chen, Yue and Hovet, Sierra and Xu, Sheng and Wood, Bradford and Ren, Hongliang and Tokuda, Junichi and Tse, Zion Tsz Ho},
  journal={International journal of computer assisted radiology and surgery},
  volume={12},
  number={11},
  pages={2023--2032},
  year={2017},
  publisher={Springer}
}

@article{taylor2017cardiovascular,
  title={Cardiovascular catheter with an expandable origami structure},
  author={Taylor, Austin J and Chen, Yue and Fok, Mable and Berman, Adam and Nilsson, Kent and Ho Tse, Zion Tsz},
  journal={Journal of medical devices},
  volume={11},
  number={3},
  pages={034505},
  year={2017},
  publisher={American Society of Mechanical Engineers}
}

@misc{origami_framework_repo,
  author       = {Anonymous},
  title        = {Origami Simulation Framework Code Repository},
  year         = {2025},
  howpublished = {\url{https://anonymous.4open.science/r/Origami-MuJoCo-Simulation-GUI/README.md}},
  note         = {Accessed: 2025-10-30}
}

@article{shi2025oribot,
  title={OriBot: a novel origami robot creation system to support children’s STEAM learning},
  author={Shi, Yan and Liu, Lijuan and Lou, Xiaolong and Lu, Yiwen and Zhang, Pan and Liu, Enmao},
  journal={Multimedia Tools and Applications},
  volume={84},
  number={26},
  pages={31723--31748},
  year={2025},
  publisher={Springer}
}

@misc{hansen2019pycma,
  author       = {Nikolaus Hansen and Youhei Akimoto and Petr Baudis},
  title        = {{CMA-ES/pycma} on {G}ithub},
  howpublished = {Zenodo, DOI:10.5281/zenodo.2559634},
  month        = feb,
  year         = 2019,
  doi          = {10.5281/zenodo.2559634},
  url          = {https://doi.org/10.5281/zenodo.2559634},
}

\end{document}